\PassOptionsToPackage{hyphens}{url}
\documentclass{article}
\usepackage{ijcai20}

\usepackage{times}
\usepackage{soul}
\usepackage[hidelinks]{hyperref}
\usepackage[utf8]{inputenc}
\usepackage[small]{caption}
\usepackage{graphicx}
\usepackage{amsmath}
\usepackage{amsthm}
\usepackage{booktabs}
\usepackage{xcolor}
\usepackage{algorithm, algpseudocode} 

\usepackage{times}
\usepackage{latexsym}
\usepackage{times}  
\usepackage{helvet} 
\usepackage{courier}  
\usepackage[hyphens]{url}  
\usepackage{arydshln}
\usepackage{epsfig}
\usepackage{subcaption}
\usepackage{amsmath}
\usepackage{amssymb}

\title{Video Question Answering on Screencast Tutorials}

\author{
Wentian Zhao$^{1*}$\and
Seokhwan Kim$^{2}$\footnote{Both authors contributed equally to this work.}\footnote{ This work was done at Adobe Research.}\and
Ning Xu$^1$\And
Hailin Jin$^1$\\
\affiliations
$^1$Adobe Research\\
$^2$Amazon Alexa AI\\
\emails
wezhao@adobe.com, 
seokhwk@amazon.com,\
\{nxu, hljin\}@adobe.com
}

\begin{document}

\maketitle

\maketitle
\begin{abstract}
  This paper presents a new video question answering task on screencast tutorials.
  We introduce a dataset including question, answer and context triples from the tutorial videos for a software.
  Unlike other video question answering works, all the answers in our dataset are grounded to the domain knowledge base. An one-shot recognition algorithm is designed to extract the visual cues, which helps enhance the performance of video question answering.
  We also propose several baseline neural network architectures based on various aspects of video contexts from the dataset.
  The experimental results demonstrate that our proposed models significantly improve the question answering performances
  by incorporating multi-modal contexts and domain knowledge.
\end{abstract}

\section{Introduction}
\label{sec:introduction}
The recent explosion of online videos on the web and social media is changing the way of transferring knowledge.
More specifically, instructional videos are getting more preferred by people to teach or learn how to accomplish a task step-by-step
and rapidly replacing conventional media mostly with written texts and few still images.
The success of video as an educational medium is largely based on its multi-modality to deliver information through visual, verbal, and even non-verbal communication at the same time in an effective and efficient manner.

Consequently, narrated instructional videos have been receiving much attention from both computer vision and natural language processing communities as useful data sources for multi-modal research.
Many studies have been conducted on various problems for instructional video understanding including procedure localization~\cite{yu2014instructional}, reference resolution~\cite{huang2017unsupervised} and visual grounding~\cite{Huang_2018_CVPR}.
On the other hand, video question answering, another major research topic based on multi-modal video understanding, has been rarely explored for instructional videos yet, despite the natural fit of the task into educational use cases.
Recently, some studies~\cite{ye2017video} have just introduced the question answering problems on instructional videos.

\begin{figure}[t]
  \includegraphics[width=\linewidth]{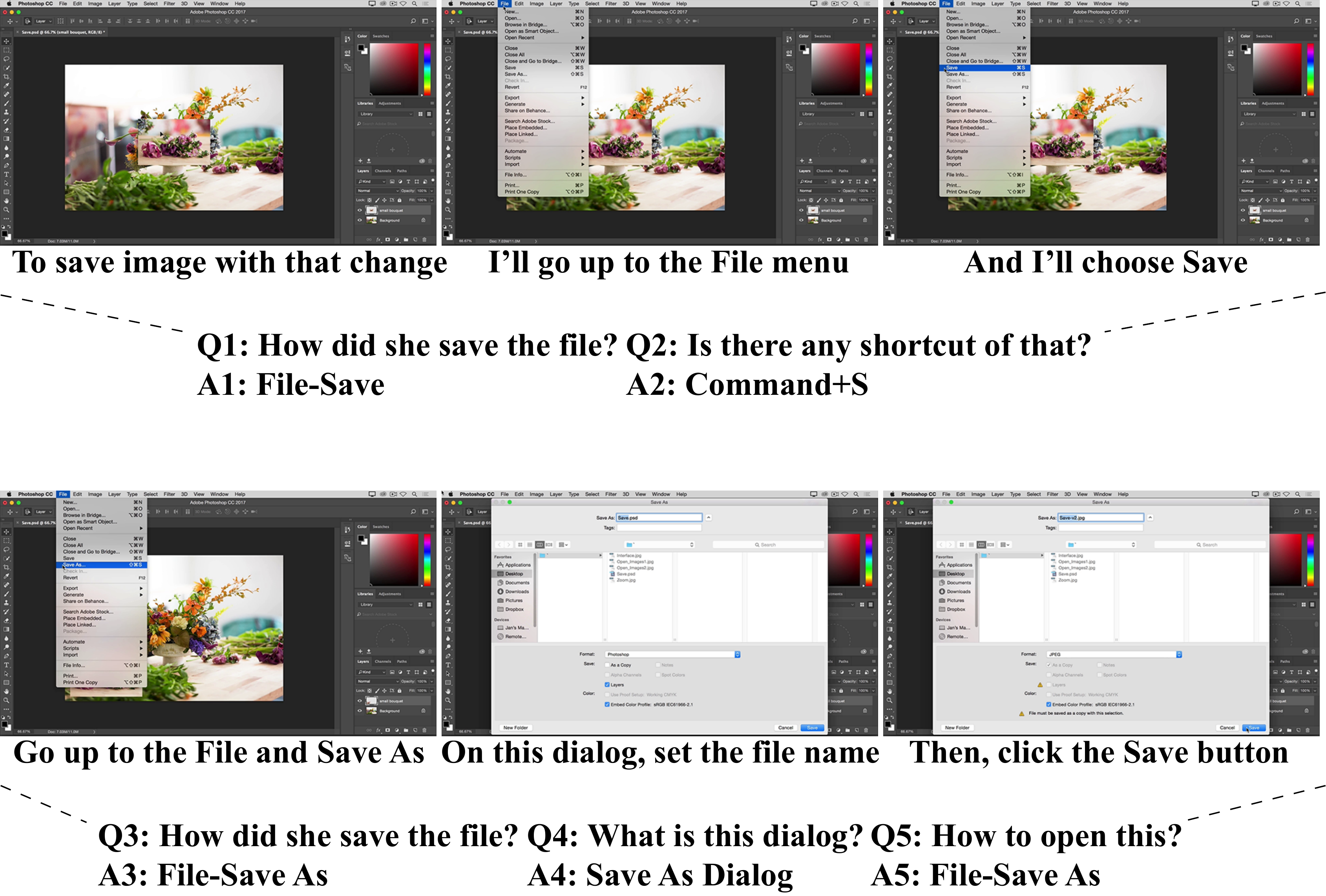}
  \caption{Examples of question answering on a screencast tutorial video for an image editing software}
  \label{fig:example}
\end{figure}

This paper presents a new instructional video question answering task on screencast tutorials which include video recordings of computer screen augmented with audio narrations to demonstrate how to use software applications.
Different from other types of instructional videos for physical real-world tasks such as cooking or do-it-yourself, screencast tutorials are mostly created and consumed in the same environment as the target application.
This aspect helps people easier to watch a screencast tutorial and follow its instructions at the same time by opening the software and the video side by side.
To develop real-time question answering capabilities in this scenario, our task is defined to take a question at anytime in the middle of a video and find the answer considering various contexts when the question is asked.
The examples in Figure~\ref{fig:example} indicate the high dependency to given contexts in selecting proper answers.
Q1 and Q3 have the same question, but the different answers from each other according to their video contexts.
The referring expressions in Q2, Q4 and Q5 also need to be resolved in a context-aware manner.
And the useful contexts are not restricted to the video contents only.
There is no explicit cue in given video contexts for answering both questions Q2 and Q5, which requires external domain knowledge.

To address the proposed task, we introduce a video question answering dataset\footnote{To download and learn more about our dataset, please see https://sites.google.com/view/pstuts-vqa/home .} collected from screencast tutorials for an image editing software.
This dataset is distinguished from the other work for the following three major characteristics.
Firstly, this dataset was collected not by automatic generation nor crowdsourcing from general public, but by the human experts of the software.
Secondly, all the questions and the answers were collected based on the localized contexts from a whole video clip with no pre-segmentation.
Above all, every answer in the dataset is grounded to its corresponding concept in a domain knowledge base.

In addition, we propose a baseline system architecture for our screencast video question answering task.
The system includes the following sub-components: text encoders for questions and transcripts, visual cue extractors from video frames, and answer encoders grounded to a domain knowledge-base.
And we compare various model configurations to fuse all the representations across different modalities to answer the questions.

The remainder of this paper is structured as follows.
Section~\ref{sec:related} compares this work with other related studies.
Section~\ref{sec:problem} presents a problem definition of our video question answering task.
Section~\ref{sec:data} introduces the question answering dataset collected on screencast tutorials for an image editing software.
Section~\ref{sec:method} describes the baseline model architectures for this problem.
Section~\ref{sec:evaluation} reports the evaluation results of these models and Section~\ref{sec:conclusions} concludes this paper.

\section{Related Work}
\label{sec:related}

\subsection{Video Question Answering}
Video question answering problems are drawing attentions of the vision community, which can be seen as an extension of image question answering.
However, the challenge of video understanding makes it an even difficult task compared with image question answering.
Many related problems have been proposed recently, while most of them are focusing on short video clips ~\cite{Unifying,spatio-temopral} with automatically generated QA pairs using some certain question generation techniques ~\cite{good-question}, of which the quality cannot be guaranteed. 
~\cite{spatio-temopral} considers the problem of open-ended video question answering from the viewpoint of spatio-temporal attentional encoder decoder learning framework. They proposed a hierarchical spatio-temporal attention network for learning the joint representation of the dynamic video contents according to the given question. To achieve spatial-temporal reasoning, \cite{multi-interaction} proposed a new attention mechanism called multi-interaction, which can capture both element-wise and segment-wise sequence interactions simultaneously. \cite{multi-step} and \cite{spatial-temporal-2} tried some other ways of interaction between spatial and temporal streams. In this work, we also explored the the effectiveness of spatial and temporal attention mechanisms in the newly proposed task, in addition to that, we further applied dual attention~\cite{kang2019dual} to model the video contexts, which is more aligned with human attention mechanism.

As a particular type of video, tutorial video is popular for video analysis in recent days, because there are tons of tutorial video resources on online platforms and the contexts are in a relatively closed environment compared to natural videos. 
Alayrac et al.~\shortcite{changing-tires-cvpr} learns the tutorial procedures in videos by leveraging the natural language annotation of the videos. 
~\cite{Youcook} proposed to learn the temporal boundaries of different steps
in a supervised manner without the aid of textual information.
To the best of our knowledge, there is no prior work on video question answering for screencast tutorials.

\subsection{Text Question Answering}
Text question answering has been extensively explored as a major research topic in the natural language processing field.
The most widely studied problem is machine reading comprehension which aims at understanding a given pair of question and source texts and generating the answer in the form of span extracted from the source texts~\cite{rajpurkar2016squad}.
In this work, we also use source texts available from the transcripts of audio narrations.
However, the machine reading comprehension methods are not applicable to our task, since many answers are not explicitly mentioned in the transcripts, but from the visual cues or external knowledge.

Another line of text question answering research has focused on answer selection problems~\cite{wang2007jeopardy,yang2015wikiqa} to find the best answer based on sentence matching between a given question and each of the answer candidates.
Our proposed task also takes the answer from a candidate pool.
But the candidate answers are not the sentences, but the concepts in a domain knowledge base, which requires different representations between questions and answers from each other.

Comparing to other knowledge-based question answering problems~\cite{berant2013semantic},
our task aims at context fusion across different modalities,
while the existing work has mainly focused on question semantics to generate the proper queries onto knowledge bases.

\section{Problem Definition}
\label{sec:problem}

We define our video question answering task as a ranking problem as follows:

\begin{equation*}
  y = {a \in A}\left(f(q, c)\cdot g(a)\right), 
\end{equation*}
where $q$ is an input question, $c$ is a given video context from either or both of video frames and transcripts, $a$ is an answer candidate from the answer pool $A$.
This work focuses on the following two main research questions:
how to fuse the multi-modal video contexts into the feature representation $f$;
and how to incorporate external domain knowledge into the answer representation $g$.

This problem formulation looks similar to the previous studies on video question answering with multiple choices~\cite{tapaswi2016movieqa,jang2017tgif,kim2017deepstory}.
However, our problem is mainly differentiated from them by taking the answer pool $A$ not from any pre-defined set for each question, but from a domain knowledge base.

\section{Data}
\label{sec:data}

\begin{table}[t]
\begin{center}
\small
\begin{tabular}{l r r}
  Entity Type & \# entities & \# options \\ \hline
  Menu & 821 & -\\
  Shortcut & 119 & -\\
  Dialog & 95 & 694\\
  Tool & 72 & 736\\
  Key & 54 & -\\
  Panel & 53 & 744\\
  Item & 15 & -\\
  Action & 7 & -\\ \hline
  Total & 1,236 & 2,196\\
\end{tabular}
\end{center}
\caption{Statistics of the domain knowledge base concepts for an image editing software}
\label{tab:stat_kb}
\end{table}

To address the proposed task, we collected a new video question answering dataset on the 76 narrated instructional videos for an image editing software. 
All the videos and their manual transcripts in English were obtained from the official website of the software.
The transcripts were preprocessed by spaCy\footnote{https://spacy.io/} for sentence segmentation and word tokenization.
Then, each sentence is aligned to its corresponding part of the video by forced alignment using Kaldi~\cite{povey2011kaldi}\footnote{ http://kaldi-asr.org/}.

To collect a high-quality question answering dataset, three image editing experts using the software were hired from UpWork\footnote{ https://www.upwork.com/}.
They were asked to watch the videos in sequence and generate a question and answer pair at a moment in the video.
Each answer is linked to an entity or an option in an existing knowledge base for the software (Table~\ref{tab:stat_kb}).
The related entity pairs are cross-linked to each other also in the knowledge base according to the relation types including 'is a', 'belongs to', 'is the shortcut of', and 'is opened by'.
For the visible entities including tools, panels, and pop-up dialogs, the knowledge base includes their example images which are used to synthesize the training data for visual cue extraction in Section~\ref{sec:Visual Cue Extraction}.

\begin{figure}[t]
  \includegraphics[width=0.37\textwidth]{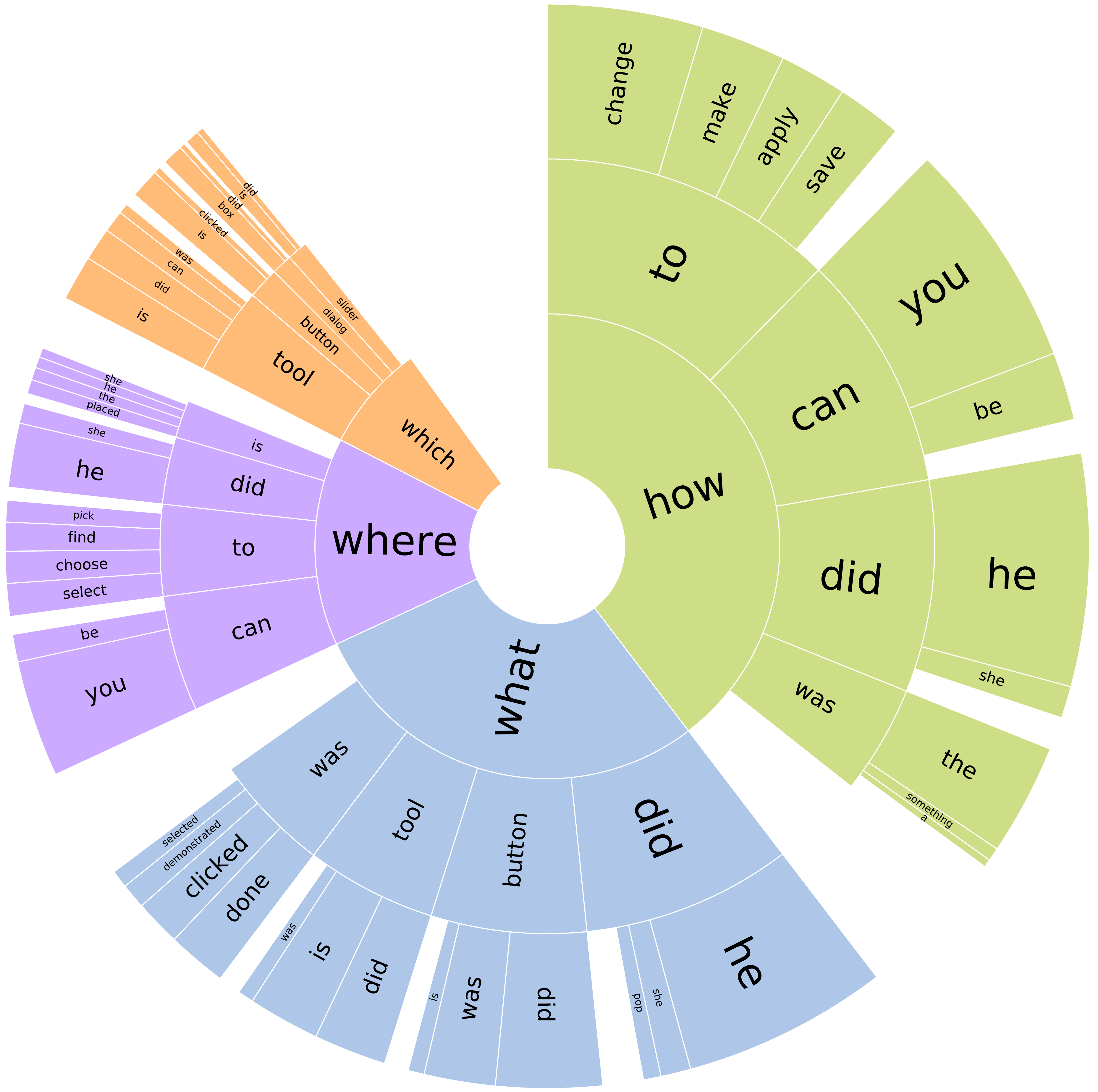}
  \centering
  \caption{Distribution of trigram prefixes of questions}
  \label{fig:prefix}
\end{figure}

\begin{figure}[t]
    \centering
  \includegraphics[width=0.70\linewidth]{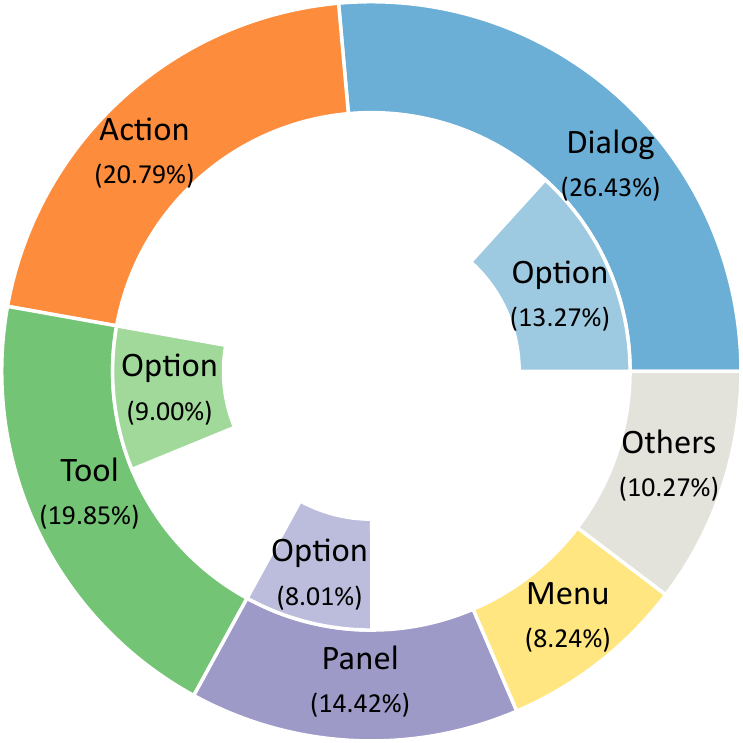}
  \caption{Distribution of answer type categories}
  \label{fig:atype}
\end{figure}

For each question $q$ and answer $a$ pair, we took the index of the transcript sentence which was being spoken when the question was asked.
Then, it is considered as the key to video context $c$ to construct an instance triple $\langle q, a, c \rangle$.
For the 2,839 unique triples collected in this first phase, we asked the experts to paraphrase each question to generate more variations.
Finally, we have 17,768 triples which were randomly divided into training, development, and test sets in Table~\ref{tab:stat_data}. As shown in the question type distributions indicated by trigram prefixes (Figure~\ref{fig:prefix}), most of the questions are about the events or the objects related to given video contexts.
69.71\% of the answers are linked to the entities of the knowledge base and the other 30.29\% of the answers are for the detailed options.
Figure~\ref{fig:atype} shows the distributions of the answer categories of the collected triples,
where 69.71\% of the instances are linked to the entities of the knowledge base and the other 30.29\% of the answers are for the detailed options.
When we match the answers to the surrounding transcripts, only 43.36\% of the instances include the exact mentions of the answers within five sentences before and after,
which largely differs from text question answering problems and implies the importance of the other contexts from visual cues and external domain knowledge.

In addition to the question and answer pairs, every video in this dataset is manually segmented for each step in the action sequence and labeled with visual cues and events which are also linked to the same knowledge-base entries as the QA instances.
These manual annotations can be utilized not only for question answering, but also for other video understanding tasks on screencast tutorials.
In this work, we leverage the ground-truth visual cue labels for training our question answering model in Section~\ref{sec:method}.

\begin{table}[t]
\begin{center}
\small
  \begin{tabular}{l r r r r}
    & \multicolumn{3}{c}{Videos} & \multicolumn{1}{c}{QAs} \\
Set & \# videos & lengths & \# sents & \# triples\\ \hline
Train & 54 & 238m & 2,660 & 12,874\\
Dev & 11 & 49m & 519 & 2,524\\
Test & 11 & 46m & 485 & 2,370 \\ \hline
Total & 76 & 333m & 3,664 & 17,768
\end{tabular}
\caption{Statistics of the datasets divided into training, development, and test purposes}
\label{tab:stat_data}
\end{center}

\end{table}

\section{Method}
\label{sec:method}
In this section, we propose a baseline model architecture for our video question answering task. 
First, we present the overview of our proposed model, then we introduce the internal components to process each information stream such as question, answer, and video contexts. 
Besides, we also illustrate some other variations in context fusion based on neural attention mechanisms.

\begin{figure*}[t]
  \includegraphics[width=\linewidth]{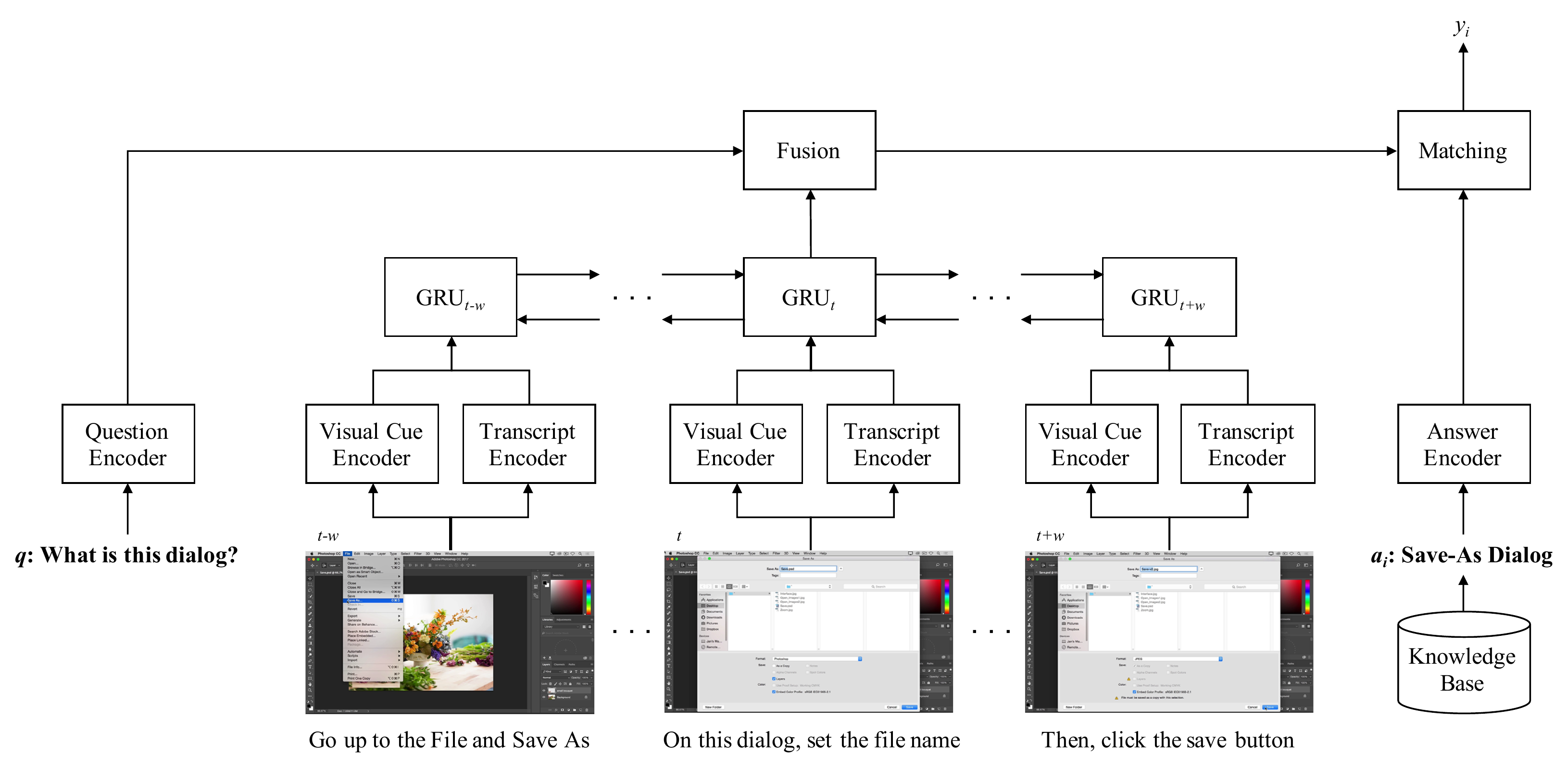}
  \caption{Base Model Architecture}
  \label{fig:architecture}
\end{figure*}

\subsection{Overall Architecture}
Figure~\ref{fig:architecture} shows the overall architecture of our baseline model.
For a question $q$ asked while the $t$-th sentence of the video transcripts is being spoken, the model takes the surrounding context $\left\{ c_{t-w}, \cdots, c_t, \cdots, c_{t+w}\right\}$, where $w$ is a window size in terms of the number of transcript sentences before and after from $t$.
Since every video segment includes both visual and language contexts, $c_j$ is defined as a pair of $v_j$ and $s_j$ which are the representations of the visual cues and the transcript sentence, respectively.

Then, the sequence from $c_{t-w}$ to $c_{t+w}$ is fed into a bidirectional recurrent layer using gated recurrent units (GRUs) (Cho et al., 2014) to learn temporal dynamics in modelling the video contexts.
From the GRU outputs, the $t$-th hidden state which is at the middle of the context sequence is taken and concatenated with the question representation of $q$.
This fused representation is forwarded to the dot product-based matching function with the answer representation of each candidate $a_i$ from the domain knowledge-base.
Finally, the candidate with the maximum matching score is selected as the answer to the question given video contexts.



\subsection{Encoders}

Our proposed model consists of the following four encoders to represent the features of questions, transcripts, visual cues, and answer candidates, respectively.

\subsubsection{Question \& Transcript Encoder}
The first type of encoder in this model aims to get the sentence representations of the question $q$ and each sentence $s_j$ in the video transcripts.
In this work, we used a common sentence encoder for both $q$ and $s_j$ based on the work by Kim~\shortcite{kim:2014:EMNLP2014}
which applies word embedding, convolution and max pooling operations in sequence.
Any other sentence representation methods can be also used for these encoders, which is out of the scope of this work.

\subsubsection{Visual Cue Encoder}
\label{sec:Visual Cue Extraction}

Software-specific visual cues play an important role in understanding the visual contexts on screencast tutorials, because most actions and operations are related to them directly. 
Therefore, we propose to extract the key visual cues for the software components including tools, panels, and pop-up dialogs first, and then use them to encode the visual contexts instead of the global video frame features as other video question answering work.

\begin{figure}[t]
  \includegraphics[width=\linewidth]{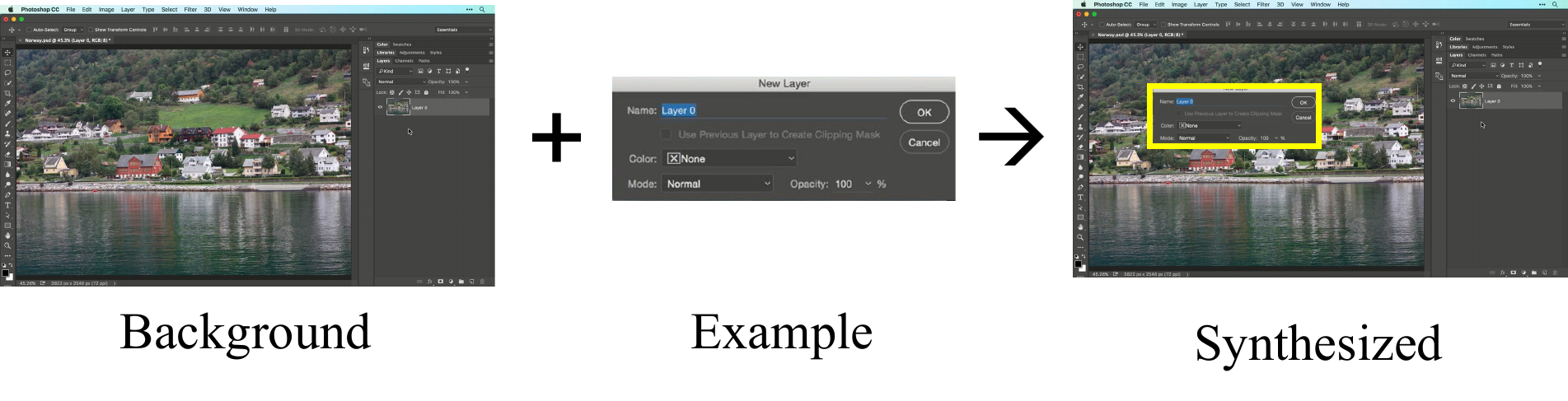}
  \caption{An example image synthesized for training visual cue detection models with no manual annotation}
  \label{fig:synthesize}
\end{figure}

The visual cue extraction procedure is divided into two parts: detection and recognition.
To detect the pop-up dialogs and panels, we train the YOLO~\cite{redmon2016you} models based on the synthetic training data without manually labeling effort.
To synthesize the images, we first collect the captured images of each target object from the domain knowledge base and then generate the images by adding pop-up items on random backgrounds as Figure~\ref{fig:synthesize}.
We adopt a similar method for tool detection of which the only difference is we do not synthesize it with backgrounds. 

With the trained models, the regions of visual cues are detected from a video frame of screencast tutorials.
Since the tool icons can be distinguished by its appearance, we build a ResNet~\cite{he2016deep} based classifier to recognize what tool is being used at each moment. 
However, panels and pop-up dialogs usually contain much more text information which makes them harder to be recognized only with the visual features.
To this end, we design a one-shot matching algorithm to recognize the panels and pop-up dialogs based on the OCR outcomes which are represented as a bag of word embedding vectors.
Algorithm~\ref{sim_computation} describes the details of our visual cue recognition method.
$sim_i$ is defined as the similarity between the i-th word in the training sample and the closest word in the test sample.
$M$ and $N$ are the number of words detected in the test sample and training sample respectively, note that 'training' and 'test' are in terms of the recognition of panels and pop-up dialogs instead of the training of question answering task. Feature matrix $\mathbf{M} \in \mathbb{R}^{N\times d}$ is the concatenation of features vectors from fastText~\cite{bojanowski2017enriching} word embeddings. $dist(\cdot,\cdot)$ is the distance between the two feature vectors, we use euclidean distance in this work. $freq(M^{i}_{test})$ is the frequency of the i-th word in the test sample.

\begin{algorithm}[bt]
\caption{Visual Cue Matching}
\label{sim_computation}
\begin{algorithmic}[1]
    \State \textbf{Initialization:} $sim_i = 0\ for\ i=1\dots M$
    \For{$i=1\dots M$}
    	\State\textbf{$min = \infty$}
    	\For{$j=1\dots N$}
        	\State\textbf{if $dist(M^{i}_{test},M^{j}_{train}) <min$:}
        	    \State\hspace{\algorithmicindent}\textbf{$min=dist(M^{i}_{test},M^{j}_{train})$}
        \EndFor
        \State $sim_i = (1/min)\cdot freq(M^{i}_{test})$
    \EndFor
    \State $similarity = \sum_{i=1}^{M} sim_i$ 
\end{algorithmic}
\end{algorithm}

The matching score is used to determine the type of each visual cue by taking the most similar entity from the knowledge base.
Each recognition result is represented also in a continuous vector space.
Since all the target entities for visual cue extraction are included in the answer candidate pool for question answering, they are finally encoded as the same representations as the corresponding answer candidates, which is described in the next subsection.

\subsubsection{Answer Encoder}
Another key to success with our proposed model architecture is how to represent each answer candidate $a_i$ into $g(a_i)$ which is matched with the fused representation $f(q,c)$.
In this work, we learn the embeddings of the answer candidates to represent each of them also as a continuous vector.
As in word representation learning, the answer embeddings can be learned from scratch or fine-tuned from pre-trained vectors.
The main difference between the two is how to initialize the answer embeddings, where the first one starts from random initialization. 
To incorporate the external domain knowledge into the latter method, we propose to fine-tune the answer embeddings from the graph embedding vectors pre-trained on the structural domain knowledge.
We convert the domain knowledge base described in Section~\ref{sec:data} into a graph structure and learn the node embeddings with DeepWalk~\cite{perozzi2014deepwalk}.
The embedding layer initialized with either random vectors or the pre-trained node embeddings is fine-tuned with the other components in the whole model architecture for question answering.

\begin{figure*}[t]
     \centering
     \begin{subfigure}[b]{0.38\linewidth}
         \centering
         \includegraphics[width=\textwidth]{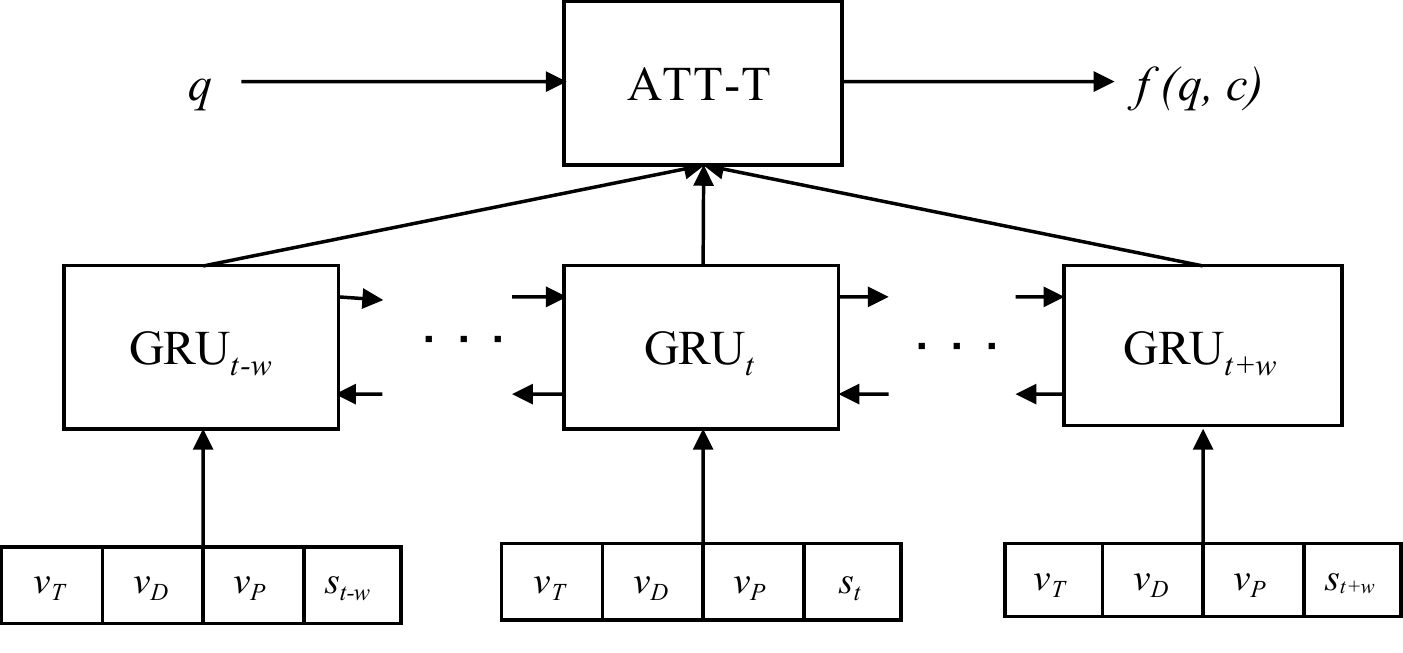}
         \caption{Temporal Attention}
         \label{fig:att_t}
     \end{subfigure}
     \hfill
     \begin{subfigure}[b]{0.23\linewidth}
         \centering
         \includegraphics[width=\textwidth]{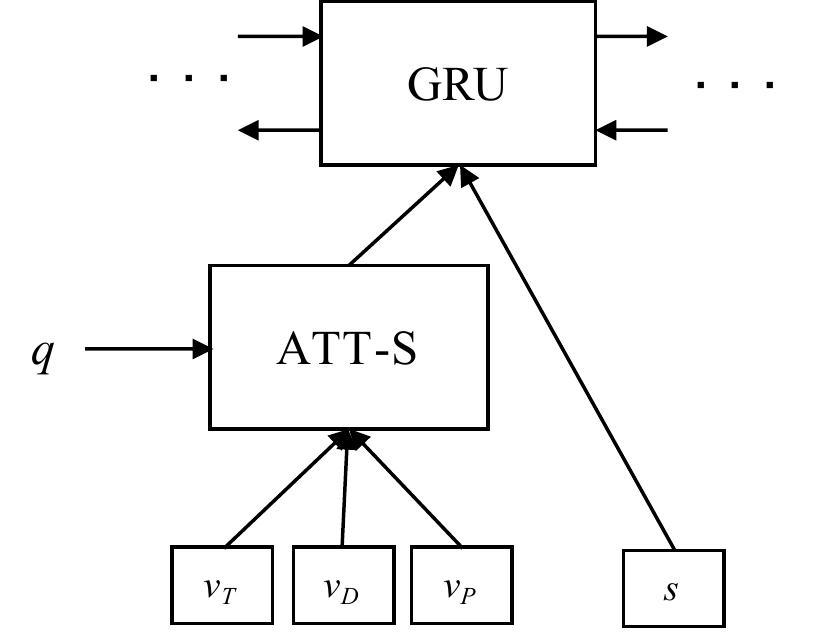}
         \caption{Spatial Attention}
         \label{fig:att_s}
     \end{subfigure}
     \hfill
     \begin{subfigure}[b]{0.30\linewidth}
         \centering
         \includegraphics[width=\textwidth]{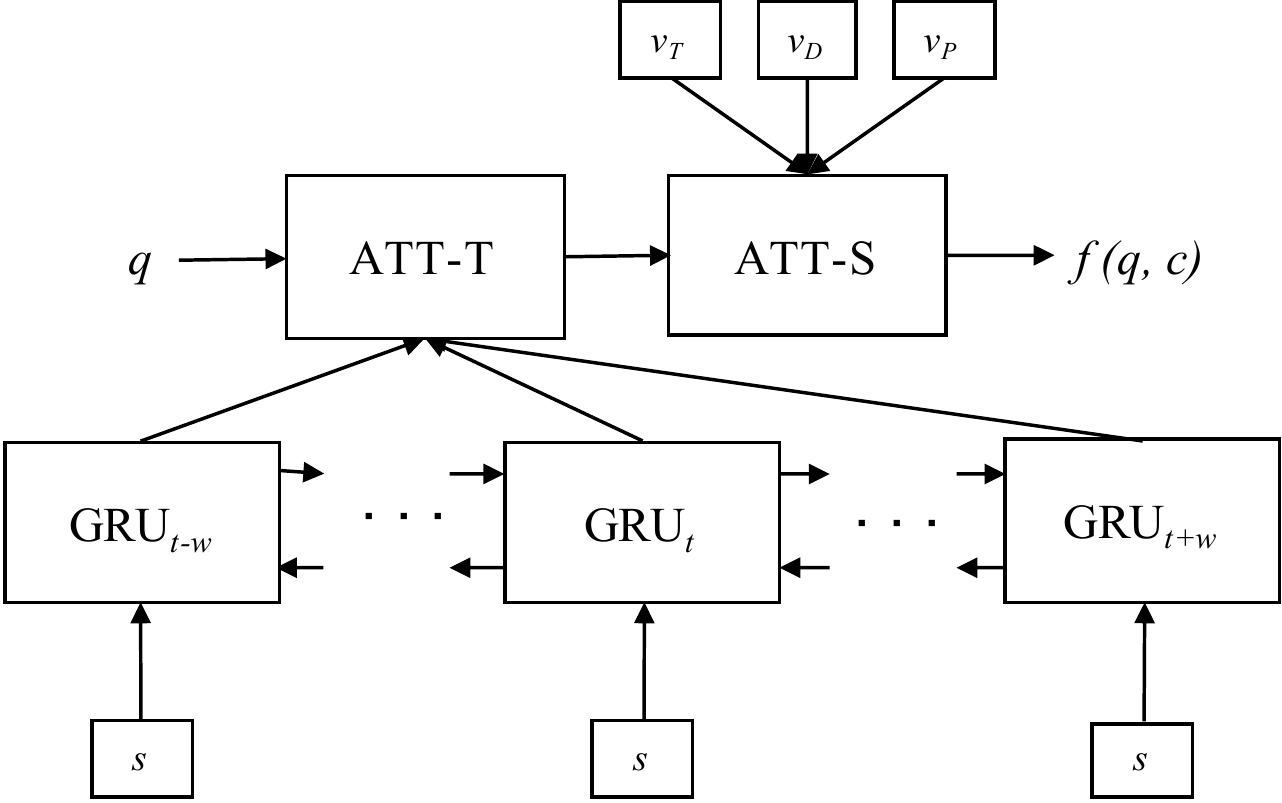}
         \caption{Dual Attention}
         \label{fig:att_d}
     \end{subfigure}
        \caption{Neural Attention Mechanisms}
        \label{fig:attentions}
\end{figure*}

\subsection{Neural Attention Mechanisms}
\label{sec:attentions}

In addition to the base model architecture, we explored further variations based on three neural attention mechanisms.
\subsubsection{Temporal Attention}
The first variation is based on temporal attentions (Figure~\ref{fig:att_t}) where the question representation is used to attend all context features at different time steps. The attention weight at each time step is computed by the softmax of MLP which takes the question and the corresponding hidden state of GRU.
Finally, we take the weighted sum of the hidden states as the video context representation.
This is a generalized version of our base model which has the hard attention only to the middle of a sequence.

\subsubsection{Spatial Attention}
Since we have multiple visual cues at each time step, we propose spatial attentions (Figure~\ref{fig:att_s}) to attend the three different visual cue streams for tools, dialogs, and panels. 
To obtain the attention weights, we apply MLP for each pair of question and visual cue representations.
Then the weighted sum of visual cue representations is concatenated with the transcript representation as the input to the bi-directional GRU.

\subsubsection{Dual Attention}
Dual attention~\cite{kang2019dual} is a way to model both temporal and spatial attentions together.
In this variation (Figure~\ref{fig:att_d}), we first apply the temporal attention on top of the GRU outputs only for the transcript sequence.
Then, the weighted sum of the transcript representations is fed into the spatial attention instead of the question itself, which is more precise than the question attended model according to the experimental results.

\section{Evaluation}
\label{sec:evaluation}

\begin{table}[]
\centering
\small
\begin{tabular}{l r r r r}
Data & Size & Precision & Recall & F1 \\ \hline
Manually labeled & 1.9k & 0.738 & 0.834 & 0.783 \\
Synthesized & 10k & 0.923 & 0.939 & 0.930 \\
\end{tabular}
\caption{Visual Cue Detection Results of Pop-up Dialog}
\label{popup_detection}
\end{table}

\begin{table}[]
\centering
\small
\begin{tabular}{l c c c c}
         & & Tools  & Dialogs & Panels \\ \hline
Accuracy & & 0.623 & 0.453  & 0.372 \\
\end{tabular}
\caption{Visual cue recognition accuracy}
\label{visual cue recog}
\end{table}

\begin{table*}[t]
\begin{center}
  \small
  \begin{tabular}{c c c c c c c c c c}
  Question & Transcript & Visual Cues & Graph Embedding & & MRR & R@1 & R@5 & R@10 & Avg Rank\\ \hline
    \checkmark & & & & & 0.4611 & 0.3460 & 0.6030 & 0.6793 & 48.12\\
    \checkmark & \checkmark & & & & 0.5610 & 0.4494 & 0.6890 & 0.7527 & 61.16\\
    \checkmark & & \checkmark & & & 0.5445 & 0.4270 & 0.6768 & 0.7527 & 68.01\\
    \checkmark & \checkmark & \checkmark & & & 0.5640 & 0.4582 & 0.6903 & 0.7688 & 38.16\\ \hdashline[.4pt/1pt]
    \checkmark & & & \checkmark & & 0.5000 & 0.3802 & 0.6451 & 0.7316 & 22.76\\ 
    \checkmark & \checkmark & & \checkmark & & 0.5832 & 0.4806 & 0.6992 & 0.7764 & \bf{18.75}\\
    \checkmark & & \checkmark & \checkmark & & 0.5886 & 0.4831 & 0.7051 & 0.7726 & 21.77\\
    \checkmark & \checkmark & \checkmark & \checkmark & & \bf{0.6637} & \bf{0.5591} & \bf{0.7869} & \bf{0.8439} & 19.27\\ \hdashline[.4pt/1pt]
    \checkmark & \checkmark & ResNet & \checkmark & & 0.5027 & 0.4013 & 0.6139 & 0.6937 & 24.60\\ 

  \end{tabular}
\caption{Comparisons of the question answering performances with different models on the test dataset.}
\label{tab:results}
\end{center}
\end{table*}

\subsection{Experimental Settings}
Based on the dataset, we first built the eight question answering models with different combinations of question, transcripts, visual cues, and graph embeddings.
All the models have the word embeddings initialized with the 300-dimensional pre-trained fastText~\cite{bojanowski2017enriching} vectors on Common Crawl dataset.
The convolutional layer in the question and transcript encoders learned 100 maps for each of three different filter sizes $\{3, 4, 5\}$.
And we set the hidden layer dimensions for GRU to 300.
For the matching component, we used dot product as a scoring function.

\subsubsection{Implementation Details}
The models were trained with Adam optimizer~\cite{DBLP:journals/corr/KingmaB14} by minimizing the negative log likelihood loss.
For training, we used mini-batch size of 128 and applied dropout on every intermediate layer with the rate of 0.5 for regularization.
The accuracy on the development set was calculated after each epoch, and then the best model was selected from the first 100 epochs for the final evaluation on the test set.

\subsubsection{Visual Cue Extraction}
For visual cue extraction, we generated 10k data samples for both panel and dialog detection training.
As shown in Table~\ref{popup_detection}, the model trained with the synthesized data outperforms the one trained on manually labeled data by a big margin.
Table~\ref{visual cue recog} shows the recognition accuracy for each visual cue type compared to the ground-truth labels on the test set videos.
All the question answering models were trained with the ground-truth visual cue labels and evaluated with the predicted outcomes by the visual cue extractors.

\subsubsection{Answer Embeddings}
For answer embeddings, we first created a domain knowledge graph including 3,432 nodes and 2,391 edges converted from the knowledge-base structure introduced in Section~\ref{sec:data}.
Then, we applied DeepWalk~\cite{perozzi2014deepwalk} algorithm based on random walk followed by skip gram~\cite{mikolov2013distributed}.
For each node in the graph, a 300-dimensional vector was trained under the same parameters used in the original work.
We initialized the answer embedding layer with these graph embedding vectors to compare with the other models with random initialization.

\subsection{Quantitative Analysis}
Table~\ref{tab:results} compares the performances of the models evaluated with the following retrieval metrics: mean reciprocal rank (MRR), recall@$k$, and the average rank of the ground truth answer,
where the higher MRR and R@$k$ scores the better results in question answering, while the lower value for the rank the higher position of the answer in the list.

Each of the contexts from transcripts and visual cues individually contributed to achieve the significantly higher performances than the models only with questions.
And the model performances were further improved when both types of the contexts were used together, which shows that these multiple modalities are complementary to each other in representing the video contexts.
In addition, the models based on pre-trained knowledge graph embeddings outperformed the other one with random initialization for every configuration,
which indicates the effectiveness of incorporating the external domain knowledge into our models.
Finally, the model with all the components achieved the best performances against the other combinations in most metrics.
Especially, this model outperformed the baseline with ResNet by large margin, which indicates the effectiveness of our proposed visual cues in video context representations. 

Table~\ref{tab:results_by_errors} shows the performances of our two best models on three subsets of the test dataset divided by the degree of prediction errors from the visual cue extractors.
The large gap between the perfect and the noisy predictions indicates that there's further room for enhancing our question answering models by improving the visual cue extraction performances, which will be one of the main action items in our future work.

\begin{table}[]
  
  \begin{center}
  \small
  
    \begin{tabular}{lrrr}
         & Wrong  & Partially correct & Correct \\ \hline
    Q+V+GE   & 0.4296 & 0.6109            & 0.6596  \\
    Q+T+V+GE & 0.5098 & 0.6766            & 0.7234 
    \end{tabular}
    \end{center}
\caption{Comparisons of the question answering performances in accuracy with different degrees of visual cue prediction errors, where Q denotes question, V is visual cues, T is transcripts and GE means Graph Embeddings.}
\label{tab:results_by_errors}
\end{table}

On top of the best base model, we applied the attention mechanisms described in Section~\ref{sec:attentions}.
Table~\ref{tab:results_attn} compares the performances with different attention mechanisms.
The models with temporal attentions failed to achieve better performances than the base models with the hard attention strategy which takes the bi-directional GRU outputs only at the time-step $t$ when each question is asked.
On the other hand, the spatial attention over the visual cues contributed to gain further improvements.
Especially, the model with dual attention mechanism achieved 1.5\% higher accuracy than the base model with no attention, which is the highest performance in this experiment.

\subsection{Qualitative Analysis}
\begin{figure}[t]
  \includegraphics[width=\linewidth]{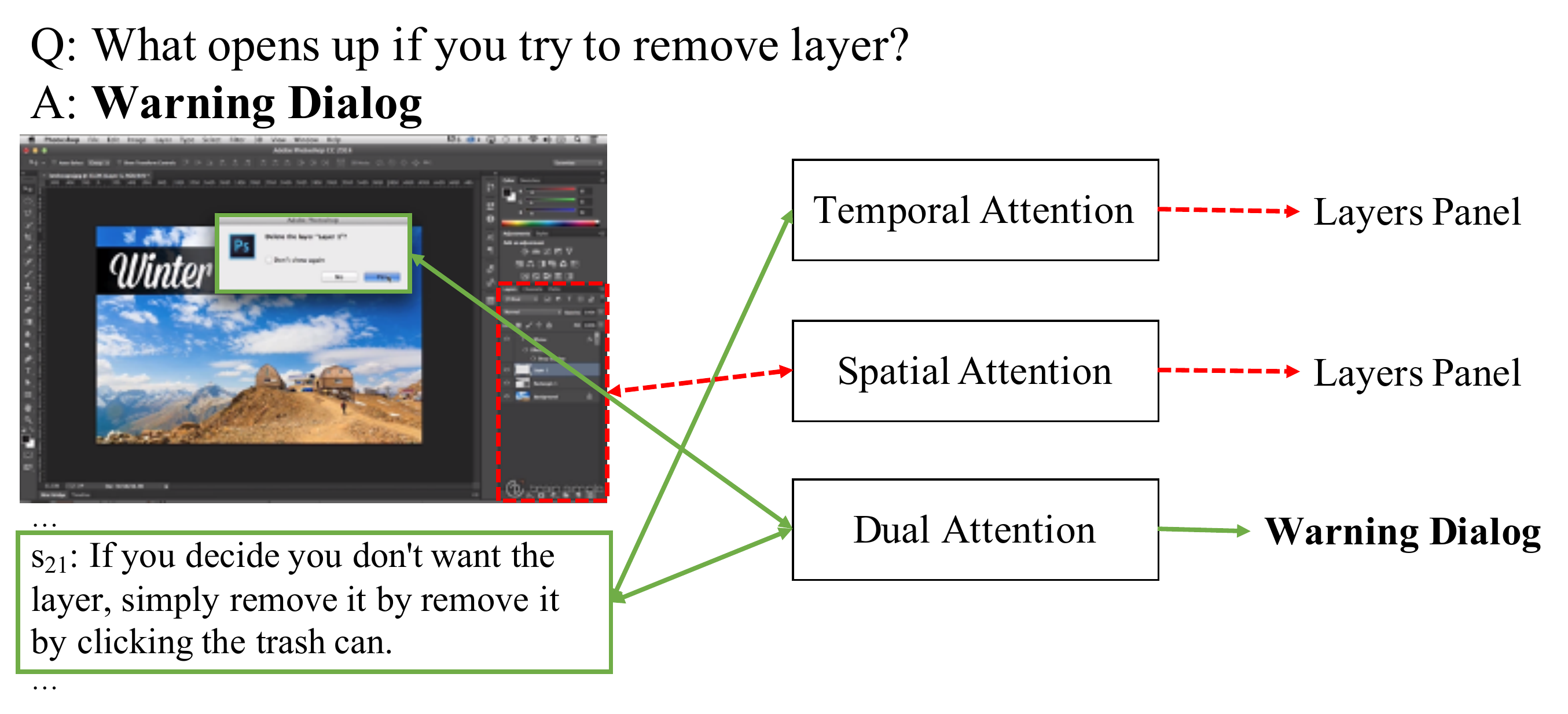}
  \includegraphics[width=\linewidth]{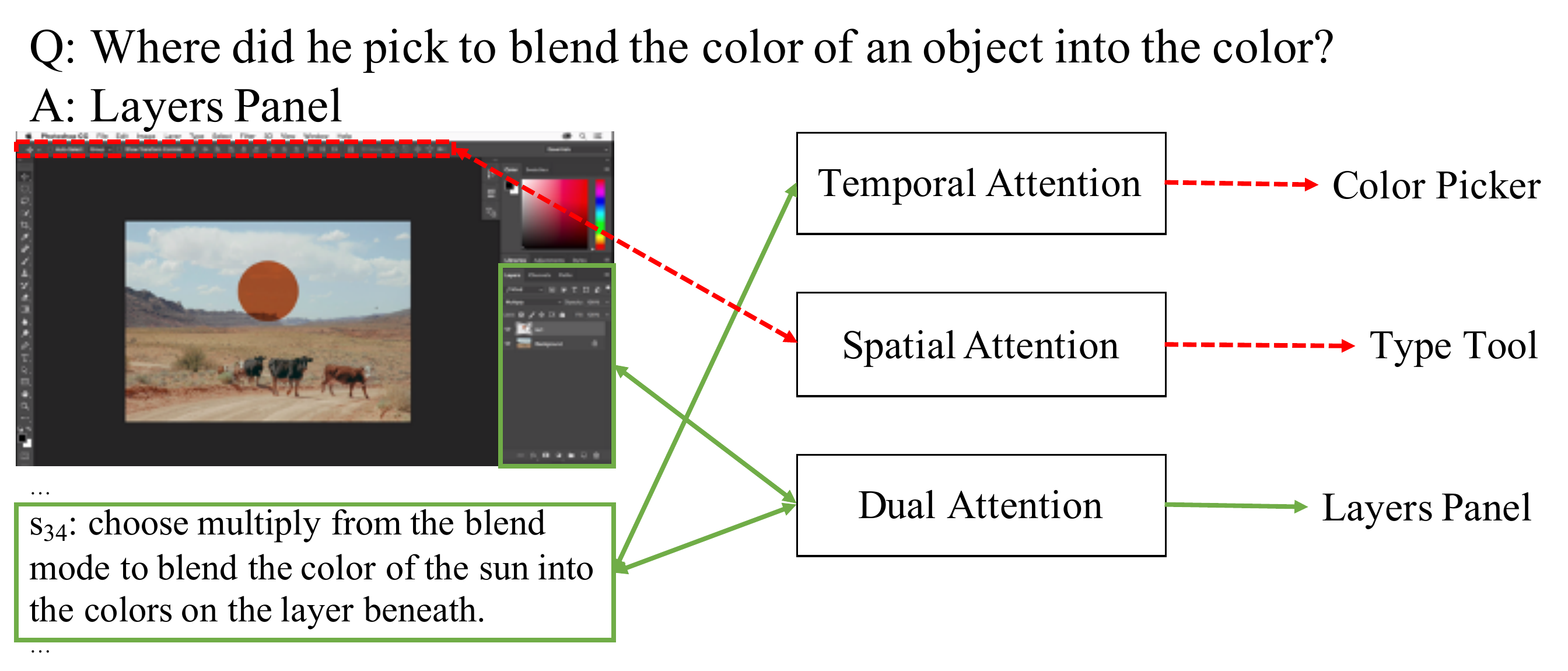}
  \caption{Visualization of the attention mechanisms. The solid green lines denote the valid attentions, while the dotted red lines show the wrong behaviors of the models.}
  \label{fig:visualization}
\end{figure}

We provide two visualization examples in Figure~\ref{fig:visualization} to illustrate in which cases the dual attention works better than spatial or temporal attentions. In these two examples, spatial attention failed to attend to the correct spatial components. Although temporal attention both attend to the correct time points, it fails to predict the answer without having spatial contexts. Dual attention is designed to attend to the spatial components based on the temporal attended contexts, which is proved to be able to alleviate this limitation.

\begin{table}[t]
  
  \begin{center}
  \small
\begin{tabular}{lrrrr}
         & No Attention & Temporal  & Spatial  & Dual \\ \hline
Accuracy & 0.5591 & 0.5414 & 0.5603 & \bf{0.5738}  
\end{tabular}
\end{center}
\caption{Comparisons of the model performances with different attention mechanisms.}
\label{tab:results_attn}
\end{table}

\section{Conclusions}
\label{sec:conclusions}

This paper presented a new video question answering task with a dataset collected in context-aware and knowledge-grounded manners on the screencast tutorial videos for a software.
Then, we proposed a neural network model architecture based on multiple encoders which represent different types of video contexts.
Experimental results showed that our proposed mechanisms to incorporate the multi-modal video contexts and the external domain knowledge helped to improve the task performances.
We also demonstrated the effectiveness of dual attention by both quantitative and qualitative analysis.


\bibliographystyle{named}
\bibliography{ijcai20}

\end{document}